\pdfoutput=1

\documentclass[11pt]{article}

\usepackage{ACL2023}

\usepackage{times}
\usepackage{latexsym}

\usepackage[T1]{fontenc}

\usepackage[utf8]{inputenc}

\usepackage{microtype}

\usepackage{inconsolata}

\usepackage{url}

\usepackage{booktabs}
\usepackage{multirow}
\usepackage{subcaption}
\usepackage{float}
\usepackage{amssymb} 
\usepackage{amsmath}

\usepackage{hyperref}
\usepackage{microtype}
\usepackage{hyperref}
\usepackage{booktabs}
\usepackage{multirow}
\usepackage{graphicx}
\usepackage{amsmath}
\usepackage{amsthm}
\usepackage{amssymb}
\usepackage{amsfonts}
\usepackage{algorithm}
\usepackage{mathtools}
\usepackage{algorithmic}
\usepackage{comment}
\usepackage{subcaption}
\usepackage{enumitem}

\usepackage{CJKutf8}

\usepackage{tikz}
\usepackage{tikz-qtree}
\usetikzlibrary{arrows,decorations.pathmorphing,backgrounds,positioning,fit,petri,shapes.misc, arrows.meta,shapes.geometric,decorations.markings,calc,shadows.blur,decorations.pathreplacing,quotes}
\definecolor{myblue}{RGB}{6, 82, 221}
\definecolor{myorange}{RGB}{211, 84, 0}
\definecolor{lowblue}{RGB}{102,178,255}
\definecolor{justblue}{RGB}{84, 160, 255}
\definecolor{mypurple}{RGB}{108, 92, 231}
\definecolor{mygray}{RGB}{158, 158, 158}
\definecolor{lowpurple}{RGB}{204,153,255}
\definecolor{lowwhite}{RGB}{255,255,255}
\definecolor{verylowpurple}{RGB}{255,102,102}
\definecolor{embcolor}{RGB}{255,255,255}
\definecolor{myred}{RGB}{235, 47, 6}
\definecolor{mygreen}{RGB}{35, 176, 56} 
\definecolor{fontgrey}{RGB}{44, 62, 80}
\definecolor{lowpurple}{RGB}{210, 180, 222}
\definecolor{mypumpkin}{RGB}{229, 152, 102}
\definecolor{lowgreen}{RGB}{171, 235, 198}
\definecolor{lowgreen2}{RGB}{186, 220, 88}
\definecolor{lowred}{RGB}{245, 183, 177}
\definecolor{lowyellow}{RGB}{241, 196, 15}
\definecolor{mypink}{RGB}{255, 118, 117}
\definecolor{bluemartina}{RGB}{18, 203, 196}
\definecolor{puffin}{RGB}{250, 152, 58}
\definecolor{grass}{RGB}{0, 148, 50}
\definecolor{cnngray}{RGB}{116, 125, 140}

\definecolor{fred}{RGB}{255, 204, 204}
\definecolor{fpurple}{RGB}{204, 229, 255}
\definecolor{fyellow}{RGB}{204, 255, 204}
\definecolor{fgreen}{RGB}{229, 204, 255}

\definecolor{flowpurple}{RGB}{150, 115, 166}
\definecolor{flowred}{RGB}{184, 84, 80}
\definecolor{flowyellow}{RGB}{214, 182, 86}
\definecolor{flowgreen}{RGB}{130, 179, 102}
\definecolor{lowlowblue}{RGB}{218, 232, 252}

\newcommand{\squishlist}{
	\begin{list}{$\bullet$}
		{ \setlength{\itemsep}{0pt}
			\setlength{\parsep}{3pt}
			\setlength{\topsep}{3pt}
			\setlength{\partopsep}{0pt}
			\setlength{\leftmargin}{1.5em}
			\setlength{\labelwidth}{1em}
			\setlength{\labelsep}{0.5em} } }

	\newcounter{Lcount}
	\newcommand{\squishlisttwo}{
		\begin{list}{\arabic{Lcount}. }
			{ \usecounter{Lcount}
				\setlength{\itemsep}{0pt}
				\setlength{\parsep}{0pt}
				\setlength{\topsep}{0pt}
				\setlength{\partopsep}{0pt}
				\setlength{\leftmargin}{2em}
				\setlength{\labelwidth}{1.5em}
				\setlength{\labelsep}{0.5em} } }
		
		\newcommand{\squishend}{
	\end{list} }

\usepackage{multirow}
\usepackage{hhline}
\usepackage{tabularx}
\usepackage{ragged2e}
\newcolumntype{Y}{>{\RaggedRight\let\newline\\\arraybackslash\hspace{0pt}}X} 
\usepackage{xcolor}
\usepackage{pgfplots, pgfplotstable}
\pgfplotsset{compat=1.17}
\usepackage{pgfpages}
\usepackage{arydshln}
\usepackage{graphicx}

\usepackage{amssymb}
\usepackage{amsfonts}

\usepackage{algorithm}
\usepackage{algorithmic}
\usepackage{pgf}
\usepackage{lipsum}
\usepackage{multicol}
\usepackage{changepage}

\usepackage[export]{adjustbox}

\usepackage[utf8]{inputenc}
 
\usepackage{amsthm}

\title{AQE: Argument Quadruplet Extraction via a Quad-Tagging Augmented Generative Approach}

\author{
\textbf{
Jia Guo\thanks{$^{*}$Equally Contributed.}~~\thanks{$^\dag$This work was done when Jia Guo was an intern at DAMO Academy, Alibaba Group.}$^\dag$\textsuperscript{\rm 1,2}~~
Liying Cheng$^{*}$\textsuperscript{\rm 1}~~
Wenxuan Zhang\textsuperscript{\rm 1}~~
Stanley Kok\textsuperscript{\rm 2}~~
Xin Li\textsuperscript{\rm 1}~~
Lidong Bing\textsuperscript{\rm 1}}\\
\textsuperscript{\rm 1}DAMO Academy, Alibaba Group~~
\textsuperscript{\rm 2}School of Computing, National University of Singapore\\
{\tt guojia@u.nus.edu}, ~~~~~{\tt skok@comp.nus.edu.sg}\\
{\tt\{liying.cheng, saike.zwx, xinting.lx, l.bing\}@alibaba-inc.com}
}

\begin{document}
\maketitle

\begin{abstract}
Argument mining involves multiple sub-tasks that automatically identify argumentative elements, such as claim detection, evidence extraction, stance classification, etc. However, each subtask alone is insufficient for a thorough understanding of the argumentative structure and reasoning process.
To learn a complete view of an argument essay and capture the interdependence among argumentative components, we need to know \textit{what} opinions people hold (i.e., claims), \textit{why} those opinions are valid (i.e., supporting evidence), 
\textit{which} source the evidence comes from (i.e., evidence type), and \textit{how} those claims react to the debating topic (i.e., stance). In this work, we for the first time propose a challenging argument quadruplet extraction task (AQE), which can provide an all-in-one extraction of four argumentative components, i.e., claims, evidence, evidence types, and stances.
To support this task, we construct a large-scale and challenging dataset. However, there is no existing method that can solve the argument quadruplet extraction. To fill this gap, we propose a novel quad-tagging augmented generative approach, which leverages a quadruplet tagging module to augment the training of the generative framework. The experimental results on our dataset demonstrate the empirical superiority of our proposed approach over several strong baselines. \footnote{Our codes and datasets are available at \url{https://github.com/guojiapub/QuadTAG}.}
\end{abstract}
\section{Introduction}

The argument plays an important role in a wide range of human activities \cite{Yuan2021OverviewOA}, from casual discussions \cite{boltuvzic2015identifying, abbott2016internet, Dusmanu2017ArgumentMO} to legal negotiations \cite{mochales2011argumentation, poudyal2017machine, Niculae2017ArgumentMW, Teruel2018IncreasingAA}, where multiple parties formulate reasons and draw conclusions.
Computational argumentation, as a growing research field, aims to automatically identify and extract the argument components presented in natural language and to predict the relationships among them \cite{Cabrio2018FiveYO}.
Given the intricate nature of the reasoning process in argumentation, identifying the various components involved and their inter-dependencies allows us to gain a deep and comprehensive understanding of the argumentative structure,
thus providing valuable information for downstream applications \cite{lawrence2019argument}.

\begin{figure}[t!]
    \center{
    \includegraphics[width=\linewidth]
    {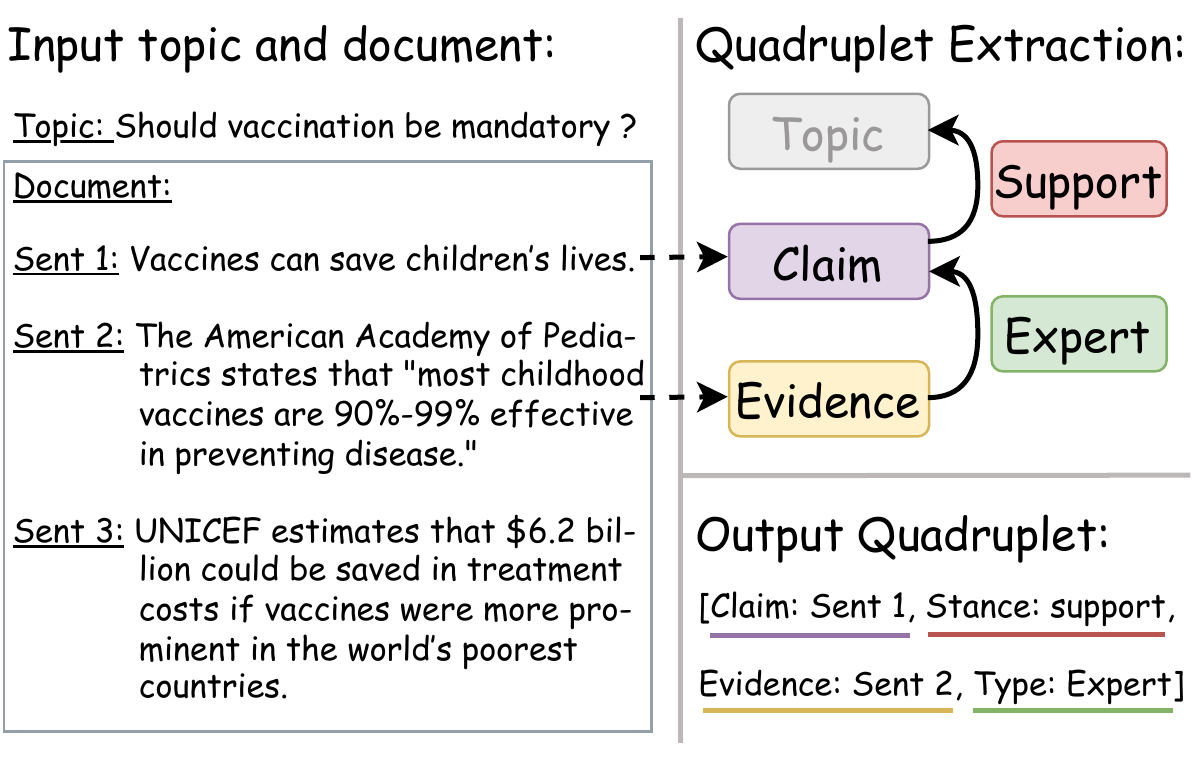}}
    \vspace{-6mm}
    \caption{\label{fig:flow} A simplified example of argument quadruplet extraction (AQE) task from our dataset. 
    Given the topic and a document containing multiple sentences, Sent 1 is a \textcolor{flowpurple}{claim} \textcolor{flowred}{supporting} the given topic, Sent 2 is a piece of \textcolor{flowgreen}{expert} \textcolor{flowyellow}{evidence} supporting the extracted claim.}
    \vspace{-3mm}
\end{figure}

Existing argument mining (AM) works focus on AM subtasks with one or a subset of the argument components, 
 such as: claim extraction \cite{aharoni2014benchmark, levy2014context},
evidence extraction \cite{rinott2015show, Singh2019ImprovingED}, 
evidence classification \cite{liga2019argumentative, afrin2020annotation}, 
stance detection \cite{Hasan2014WhyAY, bar2017stance, Hardalov2021ASO}, 
claim-evidence pair extraction \cite{cheng2022iam},
argument pair extraction \cite{Cheng2020APEAP}.
However, each of the tasks above could only provide a partial view of the whole argumentative structure, and few of them have provided a detailed analysis of the complex interplay of various components. In this work, our goal is to get a thorough understanding of the overall argumentative structures.
Hence, we propose a novel task named \textit{Argument Quadruplet Extraction} (AQE). Specifically, provided with a controversial topic and document, our AQE task aims to answer:
(1) \textit{what} opinions the party holds towards the topic (i.e., claim), 
(2) \textit{why} those opinions are tenable (i.e., evidence), 
(3) \textit{which} source the evidence comes from (i.e., evidence type),
and  
(4) \textit{how} these opinions react to the debating topic (i.e., stance). A simplified example in Figure \ref{fig:flow} illustrates the input and output of our AQE task.

To facilitate the study of this AQE task, a comprehensive dataset with all argumentative components (i.e., claim, evidence, stance, and evidence type) and their relations (i.e., claim-evidence pairing relations) is needed.
Although a previous dataset \cite{cheng2022iam} has included multiple argument elements, the evidence-type information has been largely ignored. Without knowing the attributes and source of supporting evidence, it is difficult to determine the persuasiveness and adequacy of a claim for decision-making. Moreover, claims supported by a variety of evidence types tend to be more convincing than those relying solely on one type of evidence \cite{rinott2015show}.

Therefore, we carefully formulate five evidence types based on references from relevant works \cite{addawood2016your, rinott2015show}: \texttt{Expert}, \texttt{Research}, \texttt{Case}, \texttt{Explanation}, \texttt{Others}. Our evidence types model the general way people recognize evidence and are widely applicable to various domains, such as online debates, policy reports, and academic writing. Both objective (i.e., \texttt{Research} and \texttt{Case}) and subjective (i.e., \texttt{Expert} and \texttt{Explanation}) categories of evidence are included. To ease the labeling labor,
we additionally label the type information of each piece of evidence on top of the existing IAM dataset \citep{cheng2022iam}.
The resulting comprehensive dataset is able to support our AQE task which takes a step forward to fully understand the argumentative structures and is named as \textit{Quadruplet Argument Mining} (QAM) dataset. 

Recently, the pre-trained generative models (e.g., \citealp{raffel2020exploring}) have shown effectiveness in information extraction \citep{zhang2022bias, Zhang2021TowardsGA}. However, most generative works operate at the word level and cannot learn the dependencies among sentences explicitly. To tackle the complex reasoning at the sentence level for the quadruplet extraction task, we for the first time propose a \textbf{Quad}-\textbf{T}agging \textbf{A}ugmented \textbf{G}enerative approach (QuadTAG), which leverages a novel quad-tagging approach as the augmentation module to enhance the generative framework by explicitly capturing the cross-sentence interactions for various components. The experimental results on our dataset demonstrate the effectiveness of our model over several strong baselines.

To summarize, our contributions include: 
\squishlist
    \item We propose a novel AQE task to extract a more comprehensive argument term consisting of multiple components and relations from unstructured text.
    \item To support the investigation of the proposed task, we introduce a new dataset QAM by additionally annotating the evidence types to an existing dataset.
    \item We propose an integrated generative framework augmented by a quad-tagging module for the AQE task, which can well capture the interrelations among multiple argument components. We demonstrate the empirical effectiveness on the proposed challenging QAM dataset. 
\squishend
\section{Related Work}

\subsection{Argument Mining Tasks}

\paragraph{Argument Mining Subtasks} 
As introduced earlier, there are four main elements for understanding the argument structures: \textit{what} (i.e., claims), \textit{why} (i.e., evidence), \textit{which} (i.e., types) and \textit{how} (i.e., stances).
Existing works focused on either one element or a subset of the four elements.
First, most earlier works only focused on \textit{subtask extraction}.
For instance,
\citet{levy2014context} proposed a task of context-dependent claim detection (CDCD).
In order to find the arguments supporting the extracted claims, \citet{rinott2015show} introduced the task of context-dependent evidence detection (CDED).
\citet{addawood2016your} worked on evidence classification subtask.
\citet{Hasan2014WhyAY} explored the task of stance classification.
Second, \citet{cheng2022iam} proposed a claim-evidence \textit{pair extraction} (CEPE) task.
Third, in terms of AM \textit{triplet extraction} task, researchers \cite{persing2016end, eger2017neural, ye2021end} aimed to extract claims, premises and their relations (i.e., stances) simultaneously.
In this work, we take a step further by proposing the argument \textit{quadruplet extraction} task, by incorporating the evidence type information.

\paragraph{Argumentation Analysis}
Argumentation analysis is critical to understand argumentative structures.
\citet{stab2014annotating} classified argumentative sentences into four classes: major claim, claim, premise, none.
\citet{park2014identifying} proposed the task of classifying the propositions into 3 categories: unverifiable, verifiable non-experimental, and verifiable experimental.
In this work, we focus on evidence classification, which has been shown in previous works that a claim can be supported using different types of evidence in different use cases \cite{rieke1984argumentation, seech1993writing, rieke2005argumentation}.
In social media domain, \citet{addawood2016your} classified the evidence into six types, including: news, expert, blog, picture, other, and no evidence.
For a similar data domain to our work (i.e., Wikipedia), \citet{rinott2015show} classified evidence into three categories: study, expert and anecdotal.
Inspired by the above, we further define 5 types of evidence by considering the context of claims, which includes: case, expert, research, explanation, and others.

\subsection{Argument Mining Models}

There are mainly two general types of end-to-end models for multiple AM subtasks, one is discriminative models and the other is generative models.
In terms of the discriminative models, \citet{chernodub2019targer} built a BiLSTM-CNN-CRF neural sequence tagging model to identify argumentative units and to classify them as claims or premises.
\citet{cheng2021argument} adopted a multi-task model with table-filling approach \cite{miwa2014modeling} for claim-evidence pair extraction task.
In terms of generative Models, \citet{potash2017here} applied pointer network sequence-to-sequence attention modeling for a joint argument relation extraction task and argument classification task.
\citet{bao2022AM} employed a pre-trained BART \cite{lewis2020bart} sequence-to-sequence language model with a constrained pointer mechanism (CPM) for an AM triplet extraction task. 
In this work, we aim to design a novel model with good generalization ability that is able to capture the sentence-level pairing relation explicitly by combining both discriminative and generative models.

\section{QAM Dataset}

To facilitate the study of the proposed argument quadruplet extraction (AQE) task, we create a fully annotated dataset based on the IAM dataset \cite{cheng2022iam}. We first describe the background of the original IAM dataset, followed by our data processing and human annotation details.

\subsection{The Original IAM Datset and Data Processing}

As described in \citet{cheng2022iam}, the IAM dataset is collected from English Wikipedia, which covers 123 debating topics.
This dataset is designed to support three tasks in argument mining, including claim extraction, evidence extraction, and stance classification.
Thus, it is fully labeled on the three argument components (i.e., claim, evidence, stance) and their internal relations.
In total, there are 69,666 sentences from 1,010 articles. 
4,890 claims with stances towards the given topics and 9,384 pieces of evidence together with the pairing relations of the extracted claims are labeled.
We remove some invalid sentences (e.g., only symbols or numbers) from the dataset, and eliminate those documents without any claim-evidence pair.
After the pre-processing, there are 34,369 sentences from 801 articles, with 3,407 claims and 8,319 pieces of evidence.

\subsection{Data Annotation}
\label{sec:annotation}

With the filtered dataset, we aim to further identify the specific relations between the extracted claim and evidence sentences. This enables the extended dataset to support our AQE task and highlights the critical role of evidence types in the overall argumentative structure.
The evidence type reflects how sufficiently the claims are supported. Without the evidence types, it is difficult to determine which claim is more compelling for decision-making. For example, arguments supported by evidence from research findings are more likely to be adopted in policy decisions than those that rely on subjective reasoning to support their opinions.
In the debating domain, a comprehensive speech typically incorporates various types of evidence, such as citing authoritative opinions from well-known figures or specific real-life cases. This approach enhances persuasiveness compared to relying solely on one type of evidence.
Therefore, it is a non-trivial task to understand the type information of each piece of evidence in the corpus.

We define 5 different evidence types based on previous work \citep{rinott2015show} as follows:
\squishlist
    \item \texttt{Case}: specific real-life cases, events, examples, etc.
    \item \texttt{Expert}: authoritative opinions of a professional, authority figure, scholar, official organization, etc.
    \item \texttt{Research}: results or conclusions from scientific research, statistical report, survey, etc.
    \item \texttt{Explanation}: detailed elaboration or explanation of the claim itself, reasons or impacts of the claim.
    \item \texttt{Others}: none of the above.
\squishend

To conduct the data annotation work, 4 professional data annotators are hired from a data company to annotate the type of each piece of evidence by following the annotation guidelines\footnote{More detailed annotation guidelines and examples are shown in Appendix \ref{sec:dataexamples}.}. 
The annotators are fully compensated for their work.
Each evidence sentence is independently labeled by 2 different annotators, and a third professional annotator will resolve the disagreement between the two annotators.
There are 8,392 evidence sentences annotated in total and the inter annotator agreement (IAA) is measured using Cohen’s Kappa with a value of 0.864.

\begin{table}[t!]
	\centering
	\resizebox{\linewidth}{!}{
        \begin{tabular}{lccc}
        \toprule
        Type  & \# Evidence & \% Evidence & Classification F$_1$ \\
        \midrule
        Case &  1,073 & 12.8\% & 74.26\\
        Expert &  1,538 & 18.3\% & 70.18\\
        Research  & 1,298 & 15.4\% & 77.71\\
        Explanation & 4,234 & 50.4\% & 89.78\\
        Others & 264 & 3.1\% & 27.91\\
        \bottomrule
        \end{tabular}}
    \vspace{-1mm}
    \caption{Statistics and analysis of evidence types.
    }
	\label{tab:evi_type}
	\vspace{-3mm}
\end{table}

\subsection{Data Analysis}
To examine the characteristics of our defined categories for evidence types, we conduct an exploratory analysis and train a simple RoBERTa-based sentence classifier for the claim and evidence sentences. The overall classification F$_1$ score is 81.79.
The distribution and classification performance in F$_1$ scores of each evidence type are shown in Table \ref{tab:evi_type}.
The classification performance on evidence sentences with \texttt{Explanation} types achieves a higher F$_1$ score due to sufficient data available for this type.
When comparing types of \texttt{Case}, \texttt{Expert} and \texttt{Research}, the objective types \texttt{Case} and \texttt{Research} outperform the subjective type \texttt{Expert}, despite having a relatively lower portion of quantities.

\begin{figure}[t!]
    \center{
    \scalebox{0.8}{
    \includegraphics[width=\linewidth]
    {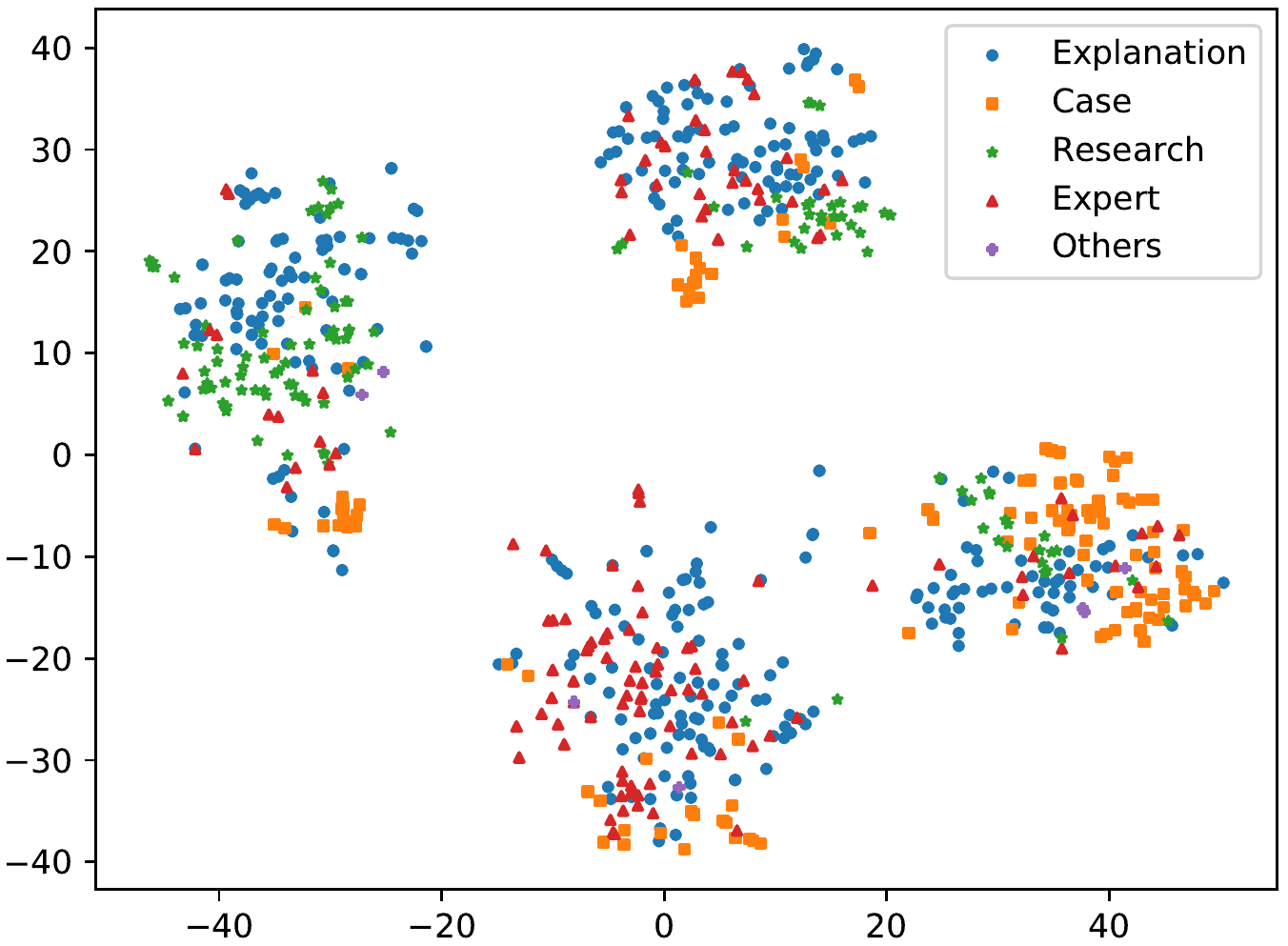}}
    }
    \vspace{-2mm}
    \caption{\label{fig:tsne} The t-SNE visualization for different evidence types across four topics.}
    \vspace{-2mm}
\end{figure}

To further analyze the properties of each evidence type, we use t-SNE algorithm \cite{Maaten2008VisualizingDU} to visualize the evidence sentences in two-dimensional space.
Specifically, we randomly select four topics that have a relatively higher amount of evidence sentences: ``\textit{Should we support family education?}'', ``\textit{Should alcohol be forbidden?}'', ``\textit{Should intellectual property rights be abolished?}'' and ``\textit{Should we fight for the Olympics?}''.
It can be observed from Figure~\ref{fig:tsne} that the distributions of evidence types vary significantly across different topics.
Furthermore, evidence sentences of types \textcolor{orange}{\texttt{Case}} and \textcolor{mygreen}{\texttt{Research}} demonstrate distinct characteristics and exhibit clear clustering within the same topic. Conversely, evidence sentences of types \textcolor{blue}{\texttt{Explanation}} and \textcolor{red}{\texttt{Expert}} show some overlap and are comparatively more challenging to differentiate.
This confirms that the evidence types pose distinct challenges, thereby indicating the highly demanding nature of performing our proposed AQE task.

\section{Task Formulation}

More formally, given a document $\mathcal{D} = [s^1, s^2, \dots, s^{n}]$ with $n$ sentences and its topic sentence $s^0$, our task aims to extract the set of argumentative quadruplets $\mathcal{Q} = \{q_k | q_k = (s^c_k, s^e_k, a_k, t_k)\}_{k=1}^{|\mathcal{Q}|}$ from the document $\mathcal{D}$, where $s^c_k, s^e_k \in \mathcal{D}$ $(c, e \in \{1, \dots, n\})$ respectively denote the claim sentence and evidence sentence. $a_k \in A$ represents the stance of the current claim sentence $s^c_k$ to the topic sentence $s^0$, $A = \{\texttt{Support}, \texttt{Against}\}$ is the set for stance labels. $t_k \in T$ denotes the evidence type for the quadruplet $q_k$. $T = \{$\texttt{Expert}, \texttt{Research}, \texttt{Case}, \texttt{Explanation}, \texttt{Others}$\}$ is the set of all evidence categories. 
\section{Model}

Distinct from existing subtasks of argument mining, our argument quadruple extraction (AQE) task brings unique challenges to current methods. It requires not only good compatibility to accommodate each argument component well but also building up the shared modeling capacities that are conducive to each subtask. The emergence of pre-trained generative model presents us with a good choice as a backbone framework to unify multiple targets into a general text-to-text learning paradigm. However, simply linearizing the argument quadruplets into a natural language sentence still can not fully exploit the underlying semantic dependencies among related components. To facilitate the task of argument quadruplet extraction, we propose an integrated framework augmented by a novel quad-tagging approach. 
\subsection{Generative Encoder}
\paragraph{Reformulated Input} Given a document $\mathcal{D} = [s^1, s^2, \dots, s^{n}]$ with $n$ sentences and its topic sentence $s^0$, sentence $s^i = [w^i_1, w^i_2, \dots, w^i_m]$ contains $m$ words. The output of AQE task requires identifying a sentence pair with the associated stance label and evidence type. However, when adapting to the text generation framework, it is inefficient to generate the original sentence of the input document during decoding especially when multiple quadruplets share the same claim or evidence sentence. To identify the sentence of interest in an efficient way and reduce the searching space of outputs, we assign each sentence with a unique symbolic ID denoted as "\texttt{\#}$i$", $(i \in [1, n])$, and insert it at the beginning of each sentence. With this symbol, we can easily recognize each sentence by its unique ID.

For our proposed quad-tagging approach, we need to obtain the hidden representation of each sentence. Inspired by the recent success of the special marker technique in information extraction \citep{atlop}, we insert two special tokens, i.e., \texttt{<SS>} and \texttt{<SE>}, at the start position and end position of the original sentence respectively, along with the symbolic ID. The contextual embedding of token \texttt{<SS>} computed by the pre-trained encoder model will be used as the sentence representation. 
\paragraph{Sentence Encoding} The reformulated input text for our proposed generative framework is defined as $\mathcal{I}(s^i) = [\texttt{<SS>}, \texttt{\#}i, w_1^i, w_2^i, \dots, w_m^i, \texttt{<SE>}]$. We concatenate the reconstructed topic sentence and all sentences in the document as long text and feed it into the T5 encoder model. The hidden representations of each input token are calculated as follows:
\begin{align}
    \centering
    \mathbf{H}_{enc} &= \text{T5\_Encoder}([\mathcal{I}(s^0), \dots, \mathcal{I}(s^n)]) ,
   \label{eq:t5_enc}
\end{align}
where $\mathbf{H}_{enc} \in \mathbb{R}^{L \times d}$ denotes the hidden representations of encoder states with length $L$ after encoding. Specifically, we use $\mathbf{h}^i_{s}$ to represent the contextual token embedding of \texttt{<SS>} for $i$-th sentence, which will be used as $i$-th sentence embedding in our proposed framework. 

\subsection{Structural Generation for Argument Quadruplet Extraction}
The straightforward way of transforming a learning task to text generation is to reformulate the expected outputs in the form of natural language sentences. However, our AQE task faces new challenges when directly adapting to text-to-text generation. As our AQE task requires identifying sentences of claim and evidence from the input document, directly generating the original text of the target sentences is space-consuming since the text can be easily retrieved from the given input document. Besides, a claim sentence is usually supported by multiple evidence sentences, repetitively generating the same claim sentence for different quadruplets will inevitably cause redundant output and a waste of computation memory. 

\begin{figure*}[t!]
    \center{
    \scalebox{1}{
    \includegraphics[width=\linewidth]
    {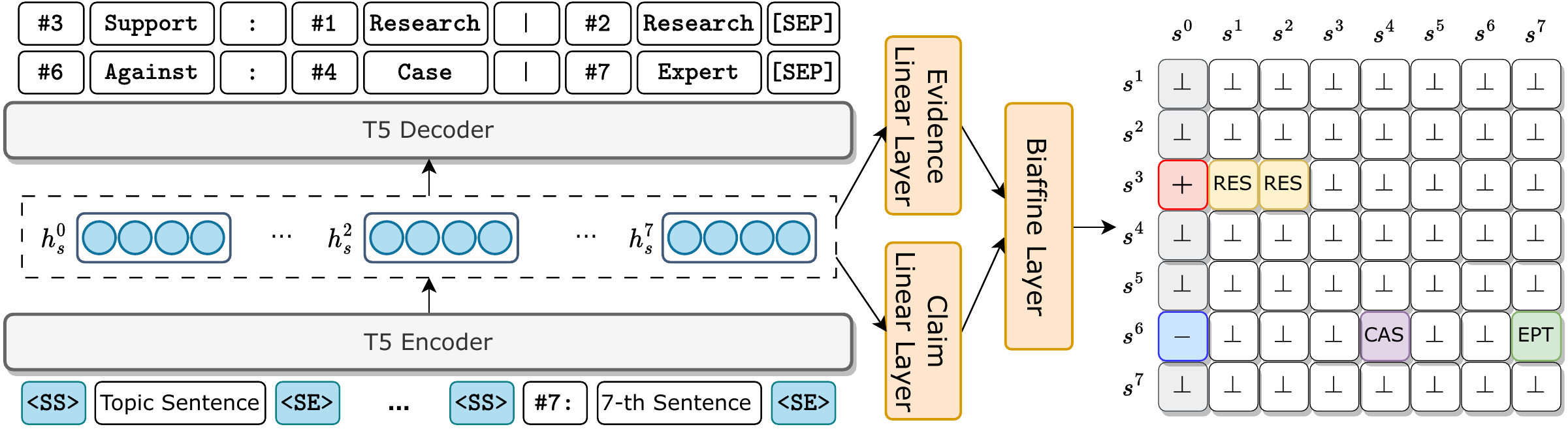}}
    }
    \vspace{-2mm}
    \caption{\label{fig:main} The overview of our proposed QuadTAG model.}
    \vspace{-2mm}
\end{figure*}

To conduct the structural generation for our AQE task in a coherent and concise way, we first define three generative templates, i.e., $\mathcal{T}_{s}, \mathcal{T}_{st}, \mathcal{T}_{et}$, for the generation outputs of target sentences $(s^c, s^e)$, stance $a$ and evidence type $t$ in a quadruplet, respectively. Concretely, $\mathcal{T}_s(s^i) = $ ``$\texttt{\#}i$'' represents the original sentence using its symbolic sentence ID. $\mathcal{T}_{st}(a)$ transforms the stance label $a \in \{\texttt{Support}, \texttt{Against}\}$ to two natural language phrases, i.e., $\mathcal{T}_{st}(\texttt{Support}) = $ ``\textit{supports the topic}'' and $\mathcal{T}_{st}(\texttt{Against}) = $ ``\textit{is against the topic}''\footnote{We also attempt another template for stance label, i.e., $\mathcal{T}'_{st}(\texttt{Support}) = $ ``\textit{positive}'' and $\mathcal{T}'_{st}(\texttt{Against}) = $ ``\textit{negative}'', please see Section \ref{sec:temp} for detailed analysis.}. We keep the original text of evidence type in the generation output, $\mathcal{T}_{et}(t) = t$, $(t \in T = \{\texttt{Expert}, \texttt{Research}, \texttt{Case}, \texttt{Explanation},$ $ \texttt{Others}\}$).
For a quadruplet $q_k = (s^c_k, s^e_k, a_k, t_k)$, we denote the expected form of its generated output as below:
\vspace{-2mm}
\begin{equation}
    \resizebox{.86\linewidth}{!}{$\begin{aligned}
    \mathcal{T}(q_k) &= \text{``}\mathcal{T}_s(s^c_k)  \ \ \mathcal{T}_{st}(a_k) \ \ \texttt{:} \ \ \mathcal{T}_s(s^e_k) \ \ \mathcal{T}_{et}(t_k)\text{''}.\label{eq:t5_dec}
\end{aligned}$}  
\end{equation}
For claims supported by multiple evidence sentences, we use the symbol ``\texttt{|}'' to concatenate different evidence and evidence types, i.e., the part of ``$\mathcal{T}_s(s^e_k) \ \ \mathcal{T}_{et}(t_k)$''. For a document with multiple claim sentences, we use a special token \texttt{[SEP]} to separate them. We provide a concrete example in the upper-left part of Figure \ref{fig:main}.

\subsection{The Quad-Tagging Augmented Module}
\label{sec:quadtag}
To facilitate the information sharing and modeling capacities for different subtasks, we propose a novel quadruplet tagging approach built in the generative backbone to explicitly enhance the interactions among sentences. For a document with $n$ sentences, we construct a table with the size of $n \times (n+1)$. Each entry has a tagging label $y_{ij}$ $(i \in [1, n], j \in [1, n+1])$. As shown in Figure \ref{fig:main}, the entries in the leftmost column of the table handle the stance detection task, i.e., $y_{i0} \in \{\perp\} \cup A$ and $\perp$ is a null label. The entries in the rest table of $n \times n$ will perform the joint tagging for the (claim, evidence, evidence type) task, i.e., $y_{ij} \in \{\perp\} \cup T$, $(j \neq 0)$. For instance, the sentence $s^3$ in Figure~\ref{fig:main} is a claim sentence and supports the topic. It is supported by two evidence sentences, i.e., $s^1$ and $s^2$, both of which belong to the \texttt{Research} type. For a non-claim sentence, such as $s^2$ in the second row, all entries in the row will be tagged with a null label ``$\perp$''.

To obtain the tagging label $y_{ij}$, we adopt a biaffine transformation layer to compute the plausibility score, which has been proven effective in related tasks \citep{biaffine}. The probability of tagging label is computed as follows: 
\begin{equation}
    \resizebox{.86\linewidth}{!}{$\begin{aligned}
    \mathbf{x}_i, \mathbf{x}_j &= \text{Linear}_c(\mathbf{h}^i_s), \text{Linear}_e(\mathbf{h}^j_{s}), \\
    P(y_{ij}) &= \text{Softmax}(\mathbf{x}_i^T\mathbf{U}\mathbf{x}_j + \mathbf{W}_i\mathbf{x}_i + \mathbf{x}_j^T\mathbf{W}_j ),\label{eq:biaffine}
\end{aligned}$}  
\end{equation}
where $\mathbf{h}^i_{s}$ and $\mathbf{h}^j_{s}$ represent the hidden representations of $i$-th and  $j$-th sentence obtained from Equation \ref{eq:t5_enc}, respectively. $\text{Linear}_e$ and $\text{Linear}_c$ are linear transformation layers for claim and evidence, respectively. $\mathbf{x}_i, \mathbf{x}_j \in \mathbb{R}^{m\times1}$ are the linearly transformed representations of the encoder outputs of claim and evidence. $\mathbf{U} \in \mathbb{R}^{m \times r \times m}$, $\mathbf{W}_i \in \mathbb{R}^{r \times m}$, $\mathbf{W}_j \in \mathbb{R}^{m \times r}$ are tunable weight parameters, $r$ is the number of all possible tags in the table and $r = |A| + |T| + 1$.

To optimize the training process, we balance the label distribution of entries with null labels by negative sampling. Specifically, $\mathcal{N}$ denotes a subset of entries randomly sampled from all entries with null labels, and $\mathcal{P}$ represents all entries with non-null labels. We conduct extensive experiments to determine the optimal ratio of negative samples, i.e., $\eta = |\mathcal{N}|/|\mathcal{P}|$, please see Appendix \ref{sec:neg-ratio} for more analysis. We adopt the cross-entropy loss function to train the quad-tagging augmented module:
\vspace{-2mm}
\begin{align}
\mathcal{L}_a = - \sum_{(i, j) \in \{ \mathcal{N} \cup \mathcal{P} \}}\sum_{k=1}^{r} y_{ij}^k\text{log}P_{\phi}(\hat{y}_{ij}^k).\label{eq:loss_tab}
\end{align}

\subsection{Training}
We finetune the pre-trained T5 model \citep{raffel2020exploring} on our QAM dataset with the autoregressive loss function shown below:
\begin{align}
\mathcal{L}_g = - \sum_{t=1}^T \text{log}P_{\theta}(y_t \ | \ \mathbf{H}_{enc}, y_{<t}), \label{eq:loss_gen}
\end{align}
where $y_t$ represents the decoder output at the $t$-th step, and $y_{<t}$ represents the previous outputs before the $t$-th step.

The final loss function for training our proposed model is defined as follows:
\begin{align}
\mathcal{L} = \mathcal{L}_g + \mathcal{L}_a. \label{eq:loss}
\end{align}
For inference, we parse the predicted quadruplets $Q'$ from the generated text sequence $y'$ by matching them with the corresponding component slots defined in the template.

\section{Experiments}
\label{sec:exp}

\subsection{Experimental Settings}

The dataset is split randomly on the document level by a ratio of 8:1:1 for training, development and testing.
The dataset statistics are shown in Table \ref{tab:stats}.
We experiment with the pre-trained RoBERTa-base model \citep{roberta} and T5-base model \citep{raffel2020exploring} for our pipeline approaches and generative methods, respectively. The max length for the output text is 512. We finetune the T5-base model on our dataset for 10 epochs with a learning rate of 1e-4 and batch size of 1. We search over $\{1,3,5,10\}$ for the number of negative examples used for the tagging loss and $\{$1e-5, 3e-5, 1e-4, 3e-4$\}$ for the learning rate. The experimental results shown in Table \ref{tab:main} are average scores and standard deviations over three runs with different random seeds. We adopt precision, recall, and F$_1$ metrics for evaluation on the development and test set. For a predicted argument quadruplet ${q_k}' = ({s^c_k}', {s^e_k}', {a_k}', {t_k}')$ to be considered correct, it has to match the ground-truth quadruplet $q_k = (s^c_k, s^e_k, a_k, t_k)$ in terms of each element and their internal relations. We run all experiments on a NVIDIA Quadro RTX 8000 GPU with 48GB GPU memory.

\subsection{Baselines}
Since there is no existing model for the argumentative quadruplet extraction task, we introduce three competitive baselines based on recent strong pre-trained language models: the pipeline approach, the pre-trained generative model, and the tagging approach. (1) The \textbf{Pipeline Approach} tackles the integrated quadruplet extraction task by decomposing it into four subtasks handled by individual pre-trained language models. The pipeline approach facilitates the information flow between tasks by utilizing the output obtained from the preceding task as the input for the subsequent task. The decomposed subtasks for the pipeline approach are claim extraction (C), stance classification (S), evidence extraction (E), and evidence type classification (T). We introduce three variants of the pipeline approach with different orders of subtasks: C-E-T-S, C-E-S-T, and C-S-E-T. The orders are determined by the basic assumption and interdependencies among the components. Specifically, the claim forms the premise for constructing an argumentative quadruple, and the remaining three components all rely on the shared claim sentence. Moreover, the evidence type relies on both the claim and evidence sentence. For the processing details of the pipeline approach, please refer to Appendix~\ref{sec:pipeline}.
(2) The \textbf{Generative Baseline} serves as a base generative model implemented on the T5-base pre-trained model \citep{raffel2020exploring}. It shares the same hyperparameter and template settings as our QuadTAG method. 
(3) The \textbf{Tagging Baseline} is the newly introduced tagging approach for our AQE task described in Section \ref{sec:quadtag}. This approach explicitly enhances the cross-sentence interactions among various components and serves as a strong discriminative baseline model. The Tagging Baseline method is trained with the encoder of the pre-trained T5-base model as the encoding backbone.

\begin{table}[t!]
	\centering
	\resizebox{0.75\linewidth}{!}{
        \begin{tabular}{lccc}
        \toprule
         \textbf{Statistics} & \textbf{Train} & \textbf{Dev} & \textbf{Test} \\
        \midrule
        \# topics & 96 & 52  & 53\\
        \# documents & 639 & 80 & 82\\
        \# paragraphs & 2,569 & 326 & 342  \\
        \# claims & 2,674 & 358 & 375 \\
        \# pieces of evidence & 6,563 & 808 & 948 \\
        \# quadruplets & 7,502 & 938 & 1,098 \\
        \bottomrule
        \end{tabular}}
    \caption{Data statistics for the QAM dataset.}
	\label{tab:stats}
\end{table}

\begin{table*}[t]
    \centering
    \resizebox{0.85\textwidth}{!}{
    \begin{tabular}{lcccccc}
    \toprule
    \multirow{2}{*}{{\textbf{Model}}} & & \textbf{Dev} & & & \textbf{Test} & \\
    \cmidrule(lr){2-4}
    \cmidrule(lr){5-7}
    & Precision & Recall & $F_1$ & Precision & Recall & $F_1$ \\
    \midrule
    Pipeline Approach (C-E-T-S) &  12.02$\pm$0.95 & 16.13$\pm$6.95 & 13.33$\pm$2.42 & 14.02$\pm$1.40 & 15.77$\pm$5.34 & 14.40$\pm$1.29 \\
    Pipeline Approach (C-E-S-T) & 11.61$\pm$0.49 & 11.73$\pm$1.67 & 11.63$\pm$0.98 & 13.47$\pm$1.35 & 11.57$\pm$1.27 & 12.44$\pm$1.23 \\
    Pipeline Approach (C-S-E-T) & \ \ 9.51$\pm$1.51 & 16.11$\pm$6.67 & 11.40$\pm$0.40 & 10.74$\pm$1.58 & 16.05$\pm$6.86 & 12.50$\pm$2.27\\
    \midrule
    Generative Baseline (T5-base) &  \underline{17.14}$\pm$2.68 & 16.60$\pm$2.58 & \underline{16.87}$\pm$2.63 & \underline{21.16}$\pm$3.55 & \underline{18.16}$\pm$2.49 & \underline{19.54}$\pm$2.94 \\
    Tagging Baseline (T5-base) & 13.98$\pm$0.89 & \textbf{18.87}$\pm$1.04 & 16.06$\pm$0.88 &  16.30$\pm$3.11 & 18.09$\pm$2.69 & 17.14$\pm$2.92\\
    \midrule
    \textbf{QuadTAG (Ours)} &  \textbf{20.55}$\pm$1.62 &  \underline{18.82}$\pm$1.66 &  \textbf{19.64}$\pm$1.65 & \textbf{24.47}$\pm$3.01 & \textbf{19.01}$\pm$1.53 & \textbf{21.39}$\pm$2.11 \\  
    \bottomrule
    \end{tabular}}
    \caption{Experimental results of our QuadTAG model and baselines for the AQE task.}
    \label{tab:main}
\end{table*}

\subsection{Main Results}
Table \ref{tab:main} shows the overall performance of our proposed QuadTAG model on the AQE task compared to the aforementioned strong baselines.
As shown in Table \ref{tab:main}, our QuadTAG model outperforms all baselines by a large margin on the F$_1$ score for both the development and test dataset. The pipeline approaches address four subtasks sequentially by separate models. We observe that both the pipeline approach (C-E-S-T) and the pipeline approach (C-S-E-T) perform worse than the pipeline approach (C-E-T-S). This is because these two approaches additionally consider the dependencies between stance and evidence type, which renders them more susceptible to the issue of error propagation.
Compared to the pipeline approaches, the end-to-end models (e.g., the generative baseline and our QuadTAG model) perform much better on three metrics. This shows that the modeling abilities developed for each subtask can be effectively transferred and leveraged for other tasks, which also implies the necessity and rationale behind the proposed AQE task in terms of empirical benefits. The tagging baseline described in Section \ref{sec:quadtag} addresses the AQE task by treating it as a classification task. However, it still falls short of both the generative baseline and our QuadTAG model, which demonstrates the generalizability of generative models for such an integrated task with multiple diverse targets involved. 
Our QuadTAG model exhibites substantial improvements of 16.4\% and 9.5\% in terms of the F$_1$ score on the development and test datasets respectively when compared to the generative baseline. The experimental results demonstrate the effectiveness of our proposed augmented module, indicating that the generative model can be further enhanced by explicitly capturing the cross-sentence interactions and the semantic inter-dependencies among distinct components. 
Both the tagging and generative baseline in Table \ref{tab:main} serve as two ablations of our QuadTAG model.

\subsection{Evaluation on Tuple and Triple Extraction}

\begin{figure}[t!]
\begin{tikzpicture}
\pgfplotsset{width=7.5cm,height=4.5cm,compat=1.8}
\begin{axis}[
    ybar,
    enlargelimits=0.3,
    legend style={at={(0.5,1.15)},
      anchor=north,legend columns=-1},
    ybar=3pt,
    ymin=10, ymax=38,
    ylabel={$F_1$ score (\%)},
    symbolic x coords={Tuple Extraction, Triple Extraction},
    xtick=data,
    xticklabel style = {font=\fontsize{9}{1}\selectfont},
    yticklabel style = {font=\fontsize{8}{1}\selectfont},
    y label style={font=\fontsize{8}{1}\selectfont},
    legend style={font=\fontsize{6}{1}\selectfont},
    nodes near coords,
    nodes near coords align={vertical},
	every node near coord/.append style={font=\fontsize{5}{1}\selectfont}
    ]

\addplot[fill=fred] coordinates {(Tuple  Extraction, 25.8) (Triple Extraction, 21.1) };
\addplot[fill=fpurple] coordinates {(Tuple Extraction, 32.3) (Triple Extraction, 27.0) };
\addplot[fill=fyellow] coordinates {(Tuple Extraction, 32.9) (Triple Extraction, 27.6) };
\addplot[fill=fgreen] coordinates {(Tuple Extraction, 34.2) (Triple Extraction, 29.2) };
\legend{Pipeline (C-E-T-S), Tagging Baseline, Generative Baseline, QuadTAG}
\end{axis}
\end{tikzpicture}
\caption{Performance comparison on the tuple extraction and triple extraction tasks.}
\label{fig:subtask}
\end{figure}
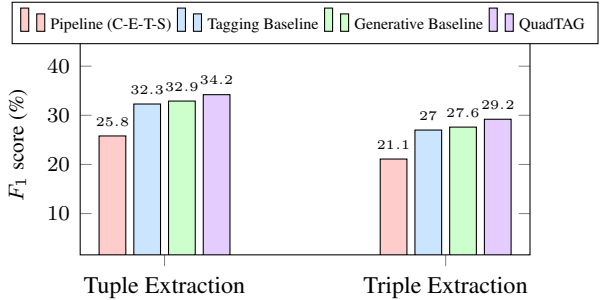

To further explore the differences in model capabilities, we present a performance comparison in Figure~\ref{fig:subtask} focusing on the extraction of a subset of argument components. Specifically, we evaluate the performance of our model and baselines in terms of extracting the (claim, evidence) tuple and the (claim, evidence, evidence type) triple.
All models are trained on the argument quadruplet dataset and evaluated on the corresponding task.
We observe that both generative models (e.g., our QuadTAG model and the generative baseline) outperform the discriminative models (e.g., the pipeline approach and the tagging baseline) for the tuple extraction and triple extraction, which further confirm the superiority of the generative framework on the complex sentence-level extraction tasks. Moreover, we observe that the tagging baseline performs comparably to the generative baseline in both tasks. This finding suggests that our proposed tagging module effectively captures the cross-sentence interactions between the claim and evidence sentences, thereby enhancing the prediction of evidence types. By harnessing the strengths of both the generative model and tagging module, our model achieves superior performance and surpasses all other models.
\subsection{Performance Breakdown on Subtasks}
We provide the performance breakdown of our model in Table~\ref{tab:break}. We evaluate our QuadTAG model on multiple subtasks at different granularities, ranging from component extraction to triple extraction. The claim component forms the basis of a quadruplet. Given that the remaining three components rely on the claim and cannot be considered alone, comparing the model performance on different joint extractions can offer valuable insights into the error distribution within the challenging AQE task. We observe that in comparison to the claim extraction, introducing the joint extraction with evidence and stance resulted in a relative decline of 37.8\% (33.08 vs. 53.20) and 26.5\% (39.12 vs. 53.20), respectively. Incorporating the extraction of evidence type, the model performance for triple extraction of (claim, evidence, evidence type) decreases by 14.9\% (28.16 vs. 33.08) compared to the tuple extraction of (claim, evidence). Furthermore, the overall performance of quadruplet extraction (i.e., 21.39 on F$_1$) is even lower than that of any of the aforementioned subtasks. The above performance degradation illustrates the challenges posed by each component and also highlights the difficulty in accurately capturing the complete quadruplet structure. 
To examine the benefit gained from integrating multiple argumentative components, we manually assign a dummy value to the argument component (e.g., we set all evidence types in the QAM dataset as \texttt{Others}), and compare the model performance with the original QuadTAG model trained on the full quadruplet dataset. From Table~\ref{tab:break}, we found that both models trained with dummy values are much worse than the original model. This further emphasizes the tight interdependence of the four components. Our quadruplet extraction can benefit subtasks by introducing other associated components and facilitating the propagation of information among them.

\begin{table}[t!]
\centering
\resizebox{\linewidth}{!}{
  \begin{tabular}{lccc}
    \toprule
    \multirow{2}{*}{{\textbf{Task}}} & & \textbf{Test}  \\
    \cmidrule(lr){2-4}
    & Precision & Recall & $F_1$  \\
    \midrule
    (Claim) & 58.94 & 48.48 & 53.20 \\
    \midrule
    (Claim, Evidence) & 37.79 & 29.42 & 33.08 \\
    (Claim, Stance) & 43.36 & 35.64 & 39.12 \\
    \midrule
    (Claim, Evidence, Evidence Type) \\
    - \textit{Trained on full quadruplets} & 32.16 & 25.05 & 28.16 \\
    - \textit{Trained on quadruplets with dummy stance} & 31.59 & 24.57 & 27.63 \\
    \midrule
     (Claim, Evidence, Stance) \\
    - \textit{Trained on full quadruplets} & 28.02 & 21.74 & 24.48 \\
    - \textit{Trained on quadruplets with dummy type} & 26.37 & 19.94 & 22.71\\    
    \bottomrule
  \end{tabular}}
   \caption{Model performance breakdown for different subtasks.}
  \label{tab:break}
\end{table}

\subsection{Generative Template Design}
\label{sec:temp}
To investigate the effects of different template designs, we evaluate the performance of our model using various templates. As shown in Table \ref{tab:template}, the prompt-based template provides some prompting words for each component, such as ``\textit{Claim Index}'' and ``\textit{Stance}''. However, it achieves poorer results than other templates, which may be due to the verbose output of the prompts, causing confusion with the original target. The order-differentiated template aims to sequentially generate four components for a quadruplet. We can observe that the empirical performance varies with different generating orders. Additionally, we offer a template with alternative textual paraphrases for the stance label, which shows the comparatively lower performance than ours. We will leave the investigation into the effects of template design for future research.

\begin{table}[!t]
    \centering
    \resizebox{\linewidth}{!}{
    \begin{tabular}{p{1.1\linewidth}ccc}
    \toprule
      \multirow{2}{*}{\textbf{Template}} & & \textbf{Test} &  \\
    \cmidrule(lr){2-4}
    & Precision & Recall & $F_1$  \\
    \midrule
\textbf{Prompt-based template} \\
 \textbf{Template:} \textit{Claim Index}: \texttt{\#[c]}, \textit{Stance}: \texttt{[a]}, \textit{Evidence Index}: \texttt{\#[e]}, \textit{Evidence Type}:  \texttt{[t]} \\ 
 \textbf{Example:} \textit{Claim Index}: \texttt{\#3}, \textit{Stance}: \texttt{positive}, \textit{Evidence Index}: \texttt{\#1}, \textit{Evidence Type}: \texttt{Research} \texttt{[SEP]} \textit{Claim Index}: \texttt{\#3}, \textit{Stance} : \texttt{positive}, \textit{Evidence Index}: \texttt{\#2}, \textit{Evidence Type}: \texttt{Research} 
  & \multirow{1}{*}{13.34} & \multirow{1}{*}{11.30} & \multirow{1}{*}{12.24} \\
    \midrule
    \textbf{Order-differentiated template} \\
 \textbf{Template:} \texttt{\#[c]}, \texttt{\#[e]}, \texttt{[t]}, \texttt{[a]} \\
 \textbf{Example:} \texttt{\#3}, \texttt{\#1}, \texttt{Research}, \textit{supports the topic} \texttt{[SEP]} \texttt{\#3}, \texttt{\#2}, \texttt{Research}, \textit{supports the topic} 
   & \multirow{1}{*}{16.11} & \multirow{1}{*}{14.29} & \multirow{1}{*}{15.15} \\
\cmidrule(lr){1-1}
 \textbf{Template:} \texttt{\#[e]}, \texttt{\#[c]}, \texttt{[a]}, \texttt{[t]}\\ \textbf{Example:} \texttt{\#1}, \texttt{\#3}, \textit{supports the topic}, \texttt{Research} \texttt{[SEP]} \texttt{\#2}, \texttt{\#3}, \textit{supports the topic}, \texttt{Research} & \multirow{1}{*}{17.65} & \multirow{1}{*}{15.35} & \multirow{1}{*}{16.42 } \\
    \midrule
\textbf{Template with other paraphrase}\\
 \textbf{Template:} \texttt{\#[c]} \texttt{[a]} : \texttt{\#[e]} \texttt{[t]} \\
 \textbf{Example:} \texttt{\#3} \texttt{positive} : \texttt{\#1} \texttt{Research} \texttt{|} \texttt{\#2} \texttt{Research} & \multirow{1}{*}{20.45} & \multirow{1}{*}{16.79} & \multirow{1}{*}{18.44} \\
    \bottomrule
    \end{tabular}}
    \caption{Experimental results of our model with different templates.}
    \label{tab:template} 
\end{table}

\section{Conclusions}
In this work, we propose a novel argument quadruplet extraction (AQE) task. To facilitate this task, we annotate a large-scale quadruplet argument mining dataset (QAM) and propose a novel quad-tagging augmented generative model (QuadTAG). Extensive experimental results and analysis validate the effectiveness of our proposed model.

\section*{Acknowledgements} 
This work is supported by Alibaba Group through Alibaba Research Intern Program. 

\section*{Limitations}
For this work, we have several limitations: first, as described in Section \ref{sec:temp}, we found that the choice of different templates and the order of generating content will both lead to performance variation. It is worthwhile to conduct a detailed investigation on this interesting problem, however, due to the limit of pages, we only experimented with limited alternative templates. Second, our proposed AQE task shares some similarities with some tasks in other domains, which means that it is possible to adapt our proposed framework to other tasks, such as relation extraction and sentiment analysis. We will leave this for future research and demonstrate its effectiveness in other domains. Last, subject to both the economic and time cost of dataset annotation, we only expand one existing dataset for our proposed AQE task. We will explore more possibilities for dataset construction for future work.

\bibliography{anthology,custom}
\bibliographystyle{acl_natbib}

\appendix
\section{Detailed Annotation Guidelines and Dataset Examples}
\label{sec:dataexamples}

\begin{table*}[ht!]
	\centering
	\resizebox{\linewidth}{!}{
        \begin{tabular}{p{0.2\linewidth} p{0.65\linewidth}cc}
        \toprule
        \textbf{Topic} & \textbf{Claim \& Evidence} & \textbf{Evidence Type} & \textbf{Stance}\\
        \midrule
        \small Should we fight for the
        & \small \textbf{Claim:} The Olympics increase valuable tourism, which can boost local economies. & \multirow{3}{*}{\small Case}  & \multirow{3}{*}{\small Support}  \\
         \cline{2-2}
         \small Olympics? &  \small \textbf{Evidence:} \colorbox{lowlowblue}{The 1984 Summer Olympics} in \colorbox{lowlowblue}{Los Angeles} netted the city a \$215 million operating surplus and \$289 million in broadcasting fees. &  &   \\
         \midrule
         
        \small Should animal testing be banned?
        & \small \textbf{Claim:} Some cosmetics and health care products must be tested on animals to ensure their safety. & \multirow{4}{*}{\small Expert} & \multirow{4}{*}{\small Contest}  \\
         \cline{2-2}
         & \small \textbf{Evidence:} \colorbox{lowlowblue}{The US Food and Drug Administration} endorses the use of animal tests on cosmetics to assure the safety of a product or ingredient. &  &   \\
         \midrule

        \small Should we ban unsustainable logging?
        & \small \textbf{Claim:} Deforestation is occurring all over the world and has been coupled with an increase in the occurrence of disease outbreaks. & \multirow{4}{*}{ \small Research} & \multirow{4}{*}{\small Support}  \\
         \cline{2-2}
         & \small \textbf{Evidence:} \colorbox{lowlowblue}{A 2017 study} in the American Economic Review \colorbox{lowlowblue}{found that} deforestation substantially increased the incidence of malaria in Nigeria. &  &   \\
         \midrule
         
        \small Should we eliminate
        & \small \textbf{Claim:} Traditional universities are a rite of passage to independent life. & \multirow{4}{*}{\small Explanation} & \multirow{4}{*}{\small Contest}  \\
         \cline{2-2}
        \small traditional universities? & \small \textbf{Evidence:} This means they have to start learning or practically using lots of skills of independent adults, such as financial management, cooking, being crime-aware, networking, and solving communication problems on their own. &  &   \\
        \bottomrule
        \end{tabular}}
    \caption{Quadruplet examples for our AQE task. Each line represents a different quadruplet with varying evidence types and stances. We highlight the signal words in the evidence sentence of different evidence types in \colorbox{lowlowblue}{blue}.
    }
	\label{tab:quad}
\end{table*}
In this section, we present our detailed annotation guidelines for human annotators. Given the topic and document information, the annotators are required to assign an evidence-type label to an evidence sentence, relying on a comprehensive comprehension of the document context and how the evidence supports its claim. As mentioned in Section~\ref{sec:annotation}, we pre-define five evidence types: \texttt{Case}, \texttt{Expert}, \texttt{Research}, \texttt{Explanation} and \texttt{Others}. We present the specific definition of each type below:

\paragraph{Case}
An evidence sentence of case type supports a claim by describing or referencing real-life cases, events, and examples to strengthen the claim. All of the following rules must be met: first, it must be an event, phenomenon, or occurrence that has taken place or existed in the real world. Second, the evidence must include at least one clearly defined and specific element related to the event, such as the individuals involved, the location, the time, and other relevant details.

The difference between this type and the explanation type is that the evidence of this type is supported by real and concrete examples, while the evidence of the explanation type remains focused on high-level analysis, reasoning, or illustration.

An argument quadruplet with case evidence is shown in the first block of Table \ref{tab:quad}.
Since the sentence clearly quotes the specific event (i.e., ``The 1984 Summer Olympics'') and the event place (i.e., ``Los Angeles''), it is considered as a real-life case to support the given claim.

\paragraph{Expert}
Expert evidence supports its claim by citing the views, comments, or suggestions of a professional, authority figure, scholar, well-known organization, official institution, representative professional group, etc.
Evidence belonging to this type can be clearly identified that the opinion or assertion in the sentence comes from a specific expert or organization, and it is essential to explicitly state the name of the expert or organization in the sentence.

Besides, we have to take note of the following: first, the difference between this type and the research type is that the evidence sentences of this type come from the viewpoints, opinions, judgments, etc. of authoritative persons or institutions, which are subjective arguments, while the evidence sentences of research type are objective arguments.
Second, if there is an overlap with the research type, it needs to be judged according to the subject of the sentence.
Third, subjective opinions, positions, judgments, and estimations from media, newspapers, publications, writings, etc., can also be labeled as the expert type.

An argument quadruplet with expert evidence is shown in the second block of Table~\ref{tab:quad}.
``The US Food and Drug Administration'' is an authoritative federal agency, and thus is labeled as expert type.

\paragraph{Research}
Evidence of the research type strengthens a claim by referencing perspectives or findings obtained from scientific research, surveys, investigation, statistical reports, or other reputable sources, including academic journals, scientific journals, etc.
At least one of the following rules must be met:
(1) The evidence sentence explicitly suggests that it pertains to a study, statistical report, or survey. Alternatively, the sentence conveys information derived from research, statistics, or surveys, typically related to research conclusions, findings, statistical data, etc. Usually, the evidence sentence of this type contains some keywords, such as ``The research shows'', ``A survey found'', ``According to the report'', etc.
(2) The evidence sentence presents a substantial amount of factual statistics or numbers derived from concrete studies, surveys, and statistics, to enhance the persuasiveness of its claim rather than relying on rough estimations.

An argument quadruplet with research evidence is shown in the third block of Table~\ref{tab:quad}.
This piece of evidence clearly states ``A 2017 study ... found that ...'', which quotes a finding of a specific study to support its claim, thus is labeled as research type.

\paragraph{Explanation}
This type of evidence enhances its claim by offering supplementary details, explanations, and elaborations on the claim sentence, as well as other relevant aspects such as the causes, consequences, and impacts associated with the claim.

An argument quadruplet with evidence of explanation type is shown in the last block of Table~\ref{tab:quad}. This evidence supports its claim by expanding upon the original assertion with more details.

\paragraph{Others}
We categorize evidence sentences that do not fit into any of the aforementioned categories as ``Others''. However, we discourage our annotators from assigning this label, as it contributes limited information about the attribute of evidence.

\paragraph{}
With the pre-defined categories, we also ask our annotators to take note of the following:
\squishlist
    \item When encountering a sentence that is difficult to decide, it is crucial to thoroughly analyze the relationship between the evidence and the claim, along with the document context, in order to determine the appropriate type.
    \item It is essential to comprehensively consider the semantic relationship between the preceding and following evidence sentences.
    \item Multiple consecutive evidence sentences can belong to different types depending on their content as well as their relationship with the claim and overall context.
\squishend

Apart from providing the above annotation guidelines, we work closely with professional data annotators whenever they have questions or they are unsure about the labels to make sure the data annotation quality.

\section{The Effect of Negative Sampling Ratio}
\label{sec:neg-ratio}
For determining the best negative ratio of the negative sampling method, we search over the range of \{1,3,5,10\}. As shown in Table \ref{tab:ns}, the model achieved the best performance when the negative ratio is 5. 
\begin{table}[htbp]
    \centering
    \resizebox{0.8\columnwidth}{!}{
    \begin{tabular}{cccc}
    \toprule
    \textbf{\# Negative Ratio} & \textbf{Precision} & \textbf{Recall} & $\textbf{F}_\textbf{1}$  \\
    \midrule
    1 & 25.11 & 18.44 & 21.27 \\
    3 & 24.62 & 18.66 & 21.23 \\
    5 & 27.91 & 20.77 & 23.81 \\
    10 & 21.97 & 17.74 & 19.63 \\
    \bottomrule
    \end{tabular}
    }
    \caption{Experimental results with different negative sampling ratios.
    }
    \label{tab:ns}
\end{table}

\section{Pipeline Processing Order}
\label{sec:pipeline}
\begin{figure}[htbp]
    \centering
    \begin{subfigure}[b]{0.3\textwidth}
           \centering
           \includegraphics[width=\textwidth]{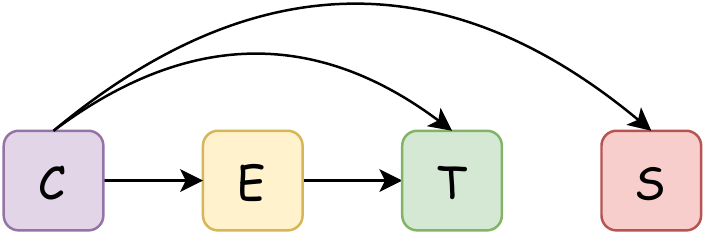}
            \caption{Pipeline approach (C-E-T-S)}
            \label{fig:a}
    \end{subfigure}
    \begin{subfigure}[b]{0.3\textwidth}
            \centering
            \includegraphics[width=\textwidth]{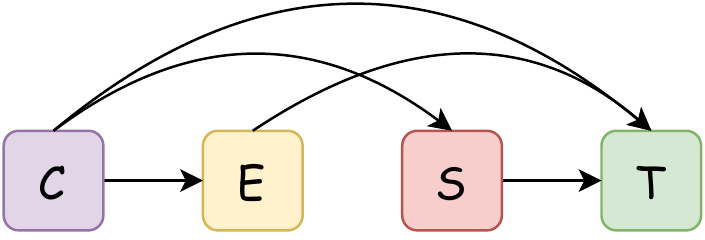}
            \caption{Pipeline approach (C-E-S-T)}
            \label{fig:b}
    \end{subfigure}
    \begin{subfigure}[b]{0.3\textwidth}
            \centering
            \includegraphics[width=\textwidth]{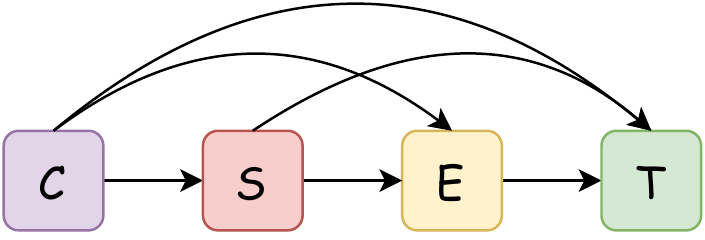}
            \caption{Pipeline approach (C-S-E-T)}
            \label{fig:c}
    \end{subfigure}
    \caption{The processing details of pipeline approaches.}
     \label{fig:pipeline}
\end{figure}

We provide the processing details in Figure~\ref{fig:pipeline} for pipeline approaches that handle four subtasks sequentially, including claim extraction (C), stance classification (S), evidence extraction (E), and evidence type classification (T). The arrow directions represent the input of each task.

\end{document}